# Software Demodulation of Weak Radio Signals using Convolutional Neural Network

Mykola Kozlenko
*Department of Information Technology*
*Vasyl Stefanyk Precarpathian National University*
Ivano-Frankivsk, Ukraine
https://orcid.org/0000-0002-2502-2447

Ihor Lazarovych
*Department of Information Technology*
*Vasyl Stefanyk Precarpathian National University*
Ivano-Frankivsk, Ukraine
https://orcid.org/0000-0001-5219-4714

Valerii Tkachuk
*Department of Information Technology*
*Vasyl Stefanyk Precarpathian National University*
Ivano-Frankivsk, Ukraine
https://orcid.org/0000-0001-7366-1676

Vira Vialkova
*Department of Cyber Security and Information Protection*
*Taras Shevchenko National University of Kyiv*
Kyiv, Ukraine
https://orcid.org/0000-0001-9109-0280

*Abstract*—In this paper we proposed the use of JT65A radio communication protocol for data exchange in wide-area monitoring systems in electric power systems. We investigated the software demodulation of the multiple frequency shift keying weak signals transmitted with JT65A communication protocol using deep convolutional neural network. We presented the demodulation performance in form of symbol and bit error rates. We focused on the interference immunity of the protocol over an additive white Gaussian noise with average signal-to-noise ratios in the range from -30 dB to 0 dB, which was obtained for the first time. We proved that the interference immunity is about 1.5 dB less than the theoretical limit of non-coherent demodulation of orthogonal MFSK signals.

*Keywords—Wide Area Monitoring, Electric Power System, Smart Grid, Weak Signal Communications, JT65A, Digital Communications, Demodulation, Frequency Shift Keying, Software Defined Radio, Machine Learning, Deep Learning, Artificial Neural Network, Deep Neural Network, Convolutional Neural Network, Interference Immunity, Symbol Error Rate, and Bit Error Rate.*

## I. Introduction

Remote monitoring in electric power systems is important to ensure the reliability and stability of the power distribution system. Traditional approaches to the data exchange in such systems usually are based on utilizing public data transmission services such as GPRS [1] and well-known others. Another popular solution is the use of the power line carrier (PLC) communications. In this case the electrical power lines are used as the transmission medium [2]. Also, the industrial communication networks such as MODBUS and some others are often used [3]. Spread spectrum signals and methods [4] are also frequently used option in wide-area power monitoring radio networks, even at nuclear power plants [5]. Frequency hopping spread spectrum (FHSS), direct-sequence spread spectrum (DSSS), chirp spread spectrum (CSS), time-hopping spread spectrum (THSS), and their combinations are commonly used as basic methods for signal processing and generation [4]. Also, stochastic communication systems based on various techniques of noise shift keying are getting popular [6]. In [7] the system is presented where the amplitude of the band-limited Gaussian noise is used for representation of binary symbols. Statistical estimates of the received signals entropy are used for information symbol demodulation and detection. In this paper we propose the use of JT65A, a highly noise immune communication protocol developed by Joe Taylor [8], for wide area monitoring. It provides possibility of data exchange with very weak signal power at the receiving point. Weak signals are very common in various long-distance digital radio communications. It is possible to use this protocol in such cases as space communications, moon-bounce communications (EME), or meteor scatter [9]. Some of mentioned approaches can be used for extra-long distance remote monitoring. The main problems for effective receiving and demodulation are the following: extremely low signal-to-noise ratio (SNR), echo delay, inter-symbol interference (ISI), time spread, Doppler Effect, libration fading, polarization, and other effects [10]. We investigated the feasibility of JT65A weak signal demodulation using convolutional artificial neural network.

## II. Background Analysis

Most early studies as well as current work focused on machine learning demodulation approaches. For instance, in the paper [11] the performance of learning-based gain is presented. It is based on the use of deep convolutional neural network for demodulation of a Rayleigh-faded wireless data signal. In the work [12] it is shown the use of such deep learning (DL) methods as deep belief network (DBN) and stacked auto-encoder (SAE) for signal demodulation in short range multi-path radio channel. The neural network-based method introduced by [13] performs automatic demodulation of various digital signals. The learning-based receiver can perform automatic selection of demodulation schema without changing receiver hardware. In the work [14] some new applications of DL for the physical layer are presented and discussed. According to the authors, a communications system is interpreted as an auto-encoder. Totally new way of representation of communications system design as an end-to-end reconstruction task was successfully established. The paper [15] is devoted to the use of spread spectrum signals and statistical entropy demodulation in telecommunication system of the household power supply meters. In the letter [16] the authors employed an artificial neural network (ANN). It is used for demodulation of optical eigenvalue modulated signals using on-off encoding. Demodulation of JT65A signals using deep dense neural network (DNN) is presented in [17]. The authors used a DNN with two hidden layers and the Softmax output activation. This paper is directly in line with the previous findings [17].

## III. Data Synthesis

Supervised machine learning approach requires a large enough amount of training data. Sometimes it is not an easy task to find sufficient amount of labeled data. In this research we have decided to use artificially synthesized dataset with the same synthesis approach as in [17]. The training set consisted of 100000 multiple frequency shift keying (MFSK) signal fragments according to [8]. Information symbols were represented by one of 64 predefined frequencies and the synchronizing tone at 1270.5 Hz. The exact frequency value can be obtained by its ordinal number as described in [8].

The duration of each symbol interval was 0.3715 second. It consisted of 4096 samples. The sample rate was 11025 samples / second. The frequency bandwidth was limited to 2500 Hz. Each symbol carries six information bits. The entire transmission contains 126 consequent time intervals. It lasts for 46.8 seconds [17]. The random phase sinusoidal wave was mixed with additive white Gaussian noise (AWGN) while synthesizing the dataset [17]. The SNR was chosen randomly within the range from -30 dB to 0 dB. The example of artificially synthesized signal waveform in the time domain at SNR = -25 dB is shown in the Fig. 1 and its autocorrelation in the Fig. 2.

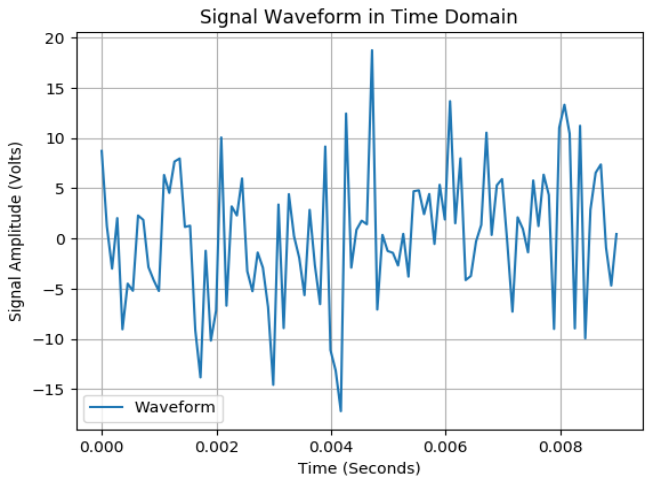

Fig. 1. Signal waveform in time domain (selected 100 successive samples) at SNR = -25 dB.

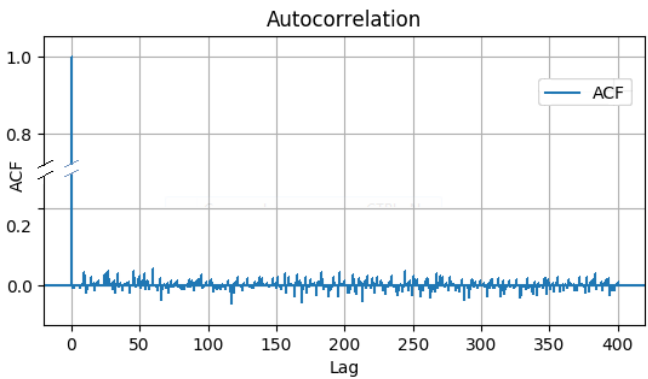

Fig. 2. Autocorrelation (selected 400 successive samples) at SNR = -25 dB.

Energy Spectral Density (ESD) of the synthesized signal was got using the Fast Fourier Transform (FFT). The FFT was calculated using the numpy.fft module as described in [17]. The NumPy is vectorized Python library for scientific computing. The ESDs for SNRs of -20 dB and -25 dB are shown in Fig. 3 and 4 correspondingly. It is obvious that at SNR of -20 dB it is possible to demodulate signals using the traditional signal processing approaches such as frequency domain filtering. It is also obvious that traditional demodulation is not possible at SNR value of -25 dB.

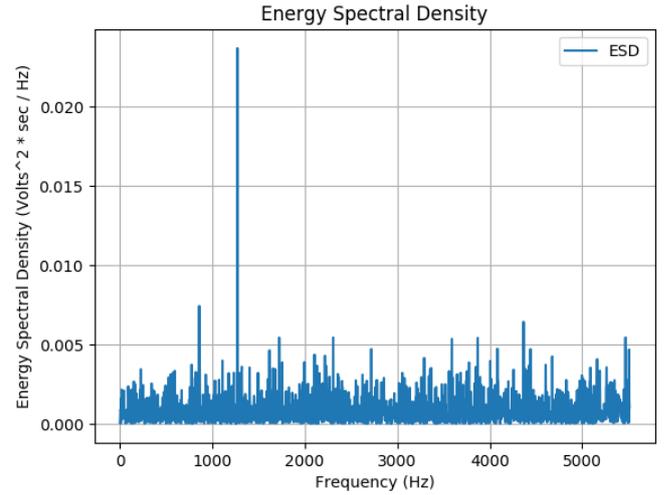

Fig. 3. Energy Spectral Density of synthesized JT65A of the synchronizing tone (1270.5 Hz) signal at SNR = -20 dB.

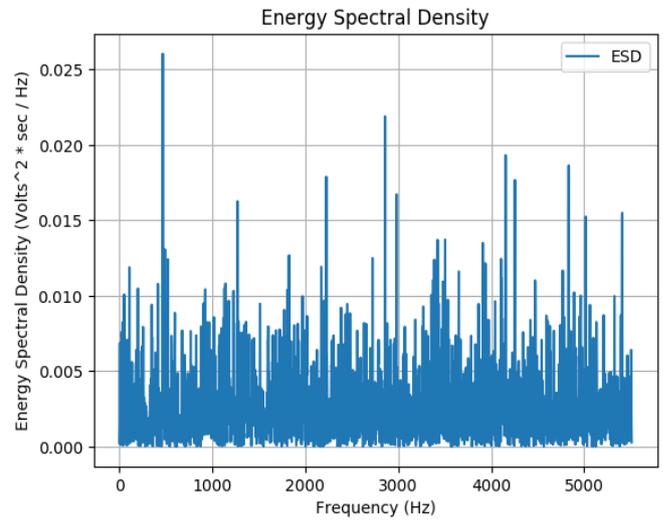

Fig. 4. Energy Spectral Density of synthesized JT65A of the synchronizing tone signal at SNR = -25 dB.

## IV. METHOD AND MODEL DESIGN

The architecture of neural network (Fig. 5) contained one one-dimensional convolution stage, one hidden dense layer with ReLU activation, and an output layer with the Softmax activation output.

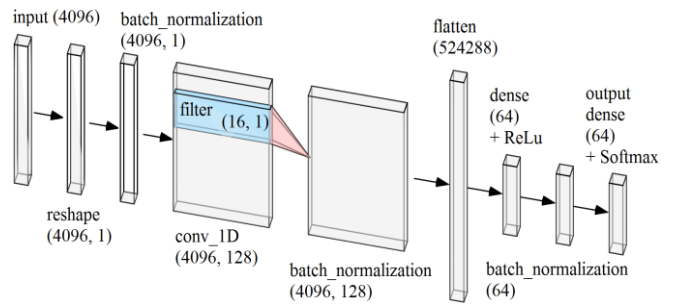

Fig. 5. Architecture of the convolutional artificial neural network model and the dimensionality of data passing through the model.

We used the TensorFlow and Keras frameworks in our work. The Tensorboard was used for visualization of training scalars and neural network structures.

There are totally 33,561,604 model parameters. There are 33,561,218 trainable parameters and 386 non-trainable ones among them.

Architecture details and hyper-parameters are as follows. Layers and their data dimensionality: input (4096), reshape (4096, 1), batch-normalization (4096, 1), one-dimension-convolution (4096, 128) with filter size (16, 1), batch-normalization (4096, 128), flatten (524288), dense (64) with ReLU activation function, batch-normalization (64), output: dense (64) with Softmax activation. Optimizer: Adam Keras built-in). The number of training epochs is 6. Loss function: categorical cross-entropy. Metrics: class-wise error rate, accuracy, precision, recall.

## V. Training and Evaluating

Training of the model was performed with the synthesized training dataset using Google Colaboratory. Training procedure takes approximately 1.4 ms per one sample (140 seconds per epoch) with GPU hardware accelerator. The number of samples per gradient update (the batch size) is 32. The training loss and accuracy against epoch number is presented in Fig. 6. Values are taken at the end of an each epoch.

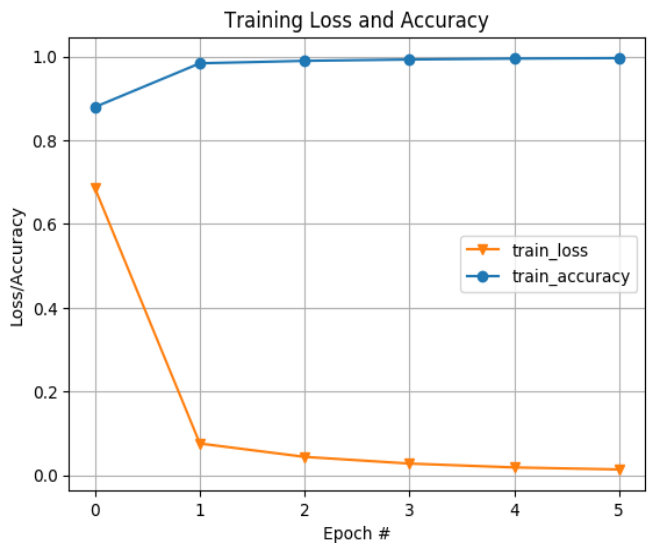

Fig. 6. Training loss and accuracy against epoch number.

We used post-predict evaluation in order to evaluate the model. Test set went through the prediction method. In our case, the test set contained 10000 symbol signals for each needed signal-to-noise ratio [17]. After that, predictions were compared to the ground truth and the confusion matrix was derived. The following class-wise and macro/micro-averaged metrics were received from the confusion matrix: error rate, accuracy, true positive rate (TPR, recall), positive predictive value (PPV, precision).

## VI. Results

Common used metric for the quality of any digital communications system is the dependence of bit error rate against normalized signal-to-noise ratio ($E_b/N_0$).

The plot of symbol error rate against SNR is presented in the Fig. 7.

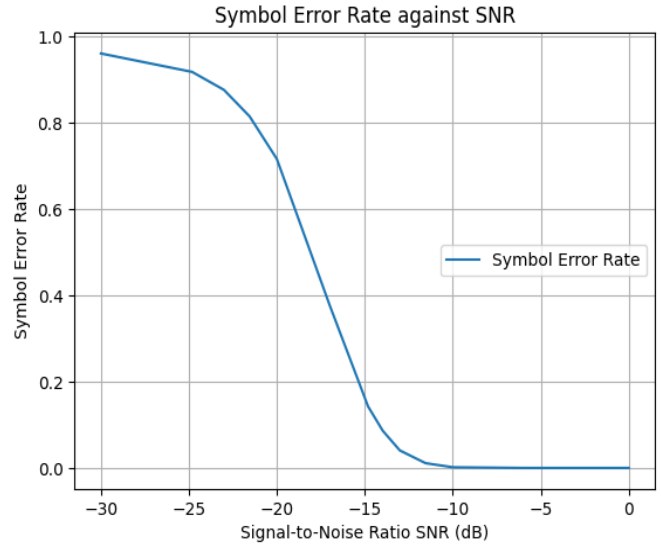

Fig. 7. Demodulation Symbol Error Rate against SNR.

Using well-known formula [4] of the determination of normalized signal-to-noise ratio given SNR we found the dependence of bit error rate (BER) against the normalized signal-to-noise ratio. It is shown in the Fig. 8.

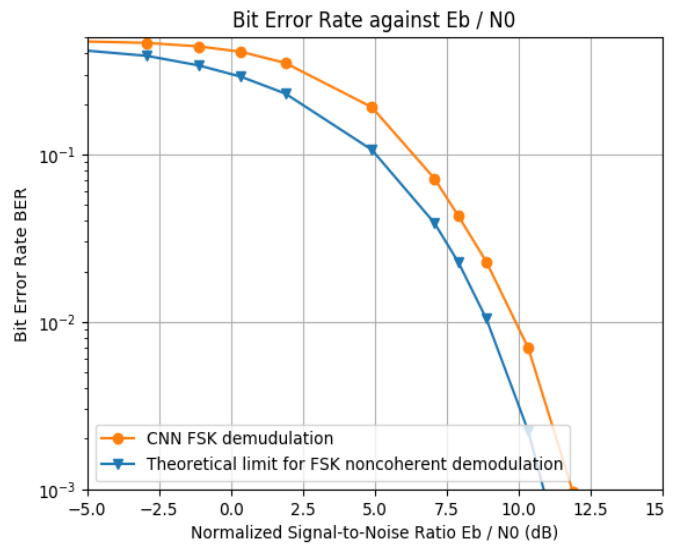

Fig. 8. Demodulation Bit Error Rate against normalized Signal-to-Noise ratio.

The theoretical limit of the BER performance for non-coherent demodulation of orthogonal FSK signals is presented in the Fig. 8 for comparison. Bit error probability for this case was found from the formula presented in [4].

BER performance (noise immunity) of the described data transmission method is less than the theoretical performance limit of non-coherent demodulation of orthogonal MFSK signals for less than 1.5 dB at bit error probability level of $10^{-2}$.

The complexity is one of most important issues related to the real time signal processing. We found the run-time complexity of the signal forward propagation by measuring the time of one data sample (one symbol interval) processing on needed hardware platform. The average processing time of one information symbol on the NVIDIA K80 GPU platform is 218 us. This does not exceed the duration of the symbol interval. As we found out, it is possible to perform real-time JT65A protocol demodulation with mentioned above hardware platform. Also, it is possible to use various other hardware platforms. Particularly, we successfully tested the following hardware platforms: Intel i5-3470 (Ubuntu Linux), Raspberry Pi 4, Samsung Galaxy J2 2018 (Android).

## VII. DISCUSSION

The overall purpose of the study was to prove the feasibility of the efficient demodulation of weak MFSK signals using convolution neural network. Our main finding suggests that the use of CNN has the acceptable outcome. CNN layers as feature extractors and dense neural layers are easy to implement computational structures with modern hardware platforms such as microcontrollers and industrial one-board microcomputers. They can be easily implemented using modern software frameworks. So, it is possible to build the software part of Software Defined Radio (SDR) equipment using this approach. As stated above, the interference immunity of the data link is high enough and approaches close to the theoretical limit of non-coherent frequency demodulation. Important advantage of the proposed method is the ability to permanent retraining on new data. This makes it easy to adapt to new conditions, signal and noise properties. The Jupyter notebooks with our experiments are open-sourced and can be downloaded from the GitHub repository [18].

## VIII. SUGGESTIONS FOR FURTHER RESEARCH

The concern about the findings of the noise immunity was the limitation to AWGN interference only. Another limitation was that artificially synthesized interference data might have different properties then real channel interferences. So, real channel data is preferable and should be recommended to use for training further models. Also, please note that we used the convolutional and dense layers only in the neural network design. The use of recurrent neural network architectures possibly can be adapted flexibly and might give new benefits. This will be a subject of our future work. We also plan to test the feasibility of the proposed method on other hardware platforms, especially on the Cortex M-series microcontrollers.

## IX. CONCLUSION

The interference immunity of JT65A communication protocol as bit error probability against normalized signal-to-noise ratio of the data exchange by MFSK signals with ML based demodulation has been obtained for the first time. It has been proved that the interference immunity is about 1.5 dB less than the theoretical limit of non-coherent demodulation of orthogonal MFSK signals. Obviously, the noise immunity of the proposed method is high enough. The method can be used as core physical technology for the data exchange in wide area monitoring systems of electric power production and distribution systems. The amount of data required to perform wide area monitoring in smart grids is very small compared with that has already been reached even for remote smart homes. Much more important is noise immunity, communication distance, stability, and reliability. Therefore, the fundamental tradeoff can be shifted toward increasing the distance of reliable communication. Thus, the proposed method is very well suited for the use in smart grids wide area monitoring systems.